\documentclass{article}

\usepackage{arxiv}

\usepackage[utf8]{inputenc} % allow utf-8 input
\usepackage[T1]{fontenc}    % use 8-bit T1 fonts
\usepackage{hyperref}       % hyperlinks
\usepackage{url}            % simple URL typesetting
\usepackage{booktabs}       % professional-quality tables
\usepackage{amsfonts}       % blackboard math symbols
\usepackage{nicefrac}       % compact symbols for 1/2, etc.
\usepackage{microtype}      % microtypography
\usepackage{lipsum}		% Can be removed after putting your text content
\usepackage{graphicx}
\usepackage{natbib}
\usepackage{doi}
\usepackage{amsmath}

\usepackage{algorithm}
\usepackage{algorithmic}
\usepackage{multirow}
\usepackage[table]{xcolor}

\title{Robust Spoofed Speech Detection via Temporal Pyramid Modeling}

\author{{\includegraphics[scale=0.06]{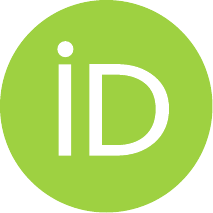}\hspace{1mm} Mahtab Masoudi Nezhad} \\
		Lane Department of Computer Science and \\
        Electrical Engineering\\
  	 West Virginia University\\
	Morgantown, WV\\\\
	\And
	\href{https://orcid.org/0000-0002-4590-7170}{\includegraphics[scale=0.06]{orcid.pdf}\hspace{1mm}Nima Karimian} \\
	Bellini College of Artificial Intelligence,\\ Cybersecurity and Computing\\
	University of South Florida\\
	Tampa, FL\\
}

\renewcommand{\shorttitle}{\textit{arXiv} Template}

\begin{document}
\maketitle

\begin{abstract}
Spoofed speech detection is increasingly challenged by realistic synthesis, voice conversion, and replay attacks, with cross-dataset generalization remaining a major limitation. This work we propose a Temporal Pyramid Adapter that utilize parallel temporal convolutions with varying receptive fields to capture multi-scale spoofing cues, ranging from local artifacts to global prosodic irregularities. We also integrated self-supervised XLS-R representations combined with front-end adapters, including Mel, Sinc, and a Temporal Pyramid design for multi-scale temporal modeling. The proposed model is evaluated cross multiple benchmark including  ASVspoof 2017, ASVspoof 2021 (DF/LA), PartialSpoof, DiffSSD, and multilingual HQ-MPSD datasets. Experimental results demonstrate that Temporal Pyramid model obtained AUC of 99.24\% and a EER of 3.87\% on the PartialSpoof database, which is significantly outperforming the base model and several SOTA baseline such as LCNN-BLSTM (9.87\% EER) and TRACE (8.08\% EER). Additionally, multilingual evaluations confirm that while spoofing artifact are independent from language. While self-supervised representations improve robustness, performance degrades under domain and language shifts, highlighting the need for better adaptation and calibration strategies. 

\end{abstract}

% keywords can be removed

\section{Introduction}

Automatic speaker verification systems are increasingly used in security-sensitive applications such as mobile authentication, banking, remote identity verification, and voice-controlled services \cite{wu2015survey,kamble2020asvspoof,tan2021survey}. These systems rely on the assumption that a speech signal contains reliable identity-related information from a genuine speaker. However, this assumption is increasingly challenged by spoofing attacks, where an adversary attempts to deceive the system using replayed speech, synthesized speech, or converted speech \cite{almutairi2022deepfake,yi2023survey}. Replay attacks use previously recorded genuine speech and play it back to the system, while text-to-speech~\cite{wu2026wildspoof} and voice conversion~\cite{kinnunen2012vulnerability} methods generate artificial speech that may resemble a target speaker. As these technologies improve, spoofed speech becomes more natural and harder to detect.

The goal of spoofed speech detection is to distinguish bona fide human speech from spoofed or manipulated audio. Early spoof detection systems often relied on handcrafted acoustic features, such as cepstral coefficients or spectral descriptors \cite{todisco2016cqcc}. Although these features can capture certain artifacts, they may fail when attacks become more diverse or when the evaluation data differs from the training data. Modern deep learning approaches have improved performance by learning discriminative features directly from audio signals, but they still face major challenges when evaluated across different datasets, attack types, recording channels, and languages.

A central difficulty in spoof detection is cross-dataset generalization. A model trained on one dataset may perform well in-domain but degrade when tested on another dataset because spoofing artifacts are not always consistent across corpora. For example, ASVspoof 2017 focuses on replay attacks \cite{kinnunen2017asvspoof}, while DiffSSD focuses on synthetic speech from modern text-to-speech systems \cite{diffssd2024}. PartialSpoof introduces localized manipulations where only part of an utterance is spoofed, making detection more difficult than full-utterance classification \cite{partialspoof2021,partialspoof2023}. ASVspoof 2021 DF and LA further introduce large-scale evaluation conditions \cite{asvspoof2021} that differ from older benchmark datasets. Because of these differences, a robust system must learn representations that capture general spoofing cues rather than dataset-specific artifacts.

This work investigates using XLS-R, a large-scale self-supervised speech representation model \cite{xlsr2022}. Self-supervised speech models are trained on large amounts of unlabeled speech and can learn general acoustic and linguistic representations. However, their performance for spoof detection can depend strongly on how the input waveform is represented. Therefore, we evaluate several front-end adapters before the XLS-R backbone. The base configuration passes the waveform directly to the model. The Mel adapter introduces a time-frequency representation. The Sinc adapter applies learnable frequency-aware filters. The Temporal Pyramid adapter uses parallel temporal convolutions with different receptive fields to capture multi-scale spoofing cues. This adapter-based framework is evaluated under cross-dataset conditions, including training on ASVspoof 2017 and testing on ASVspoof 2021 DF and LA, as well as training on PartialSpoof and testing on ASVspoof 2021 and DiffSSD.

Beyond the challenge of language shift itself, a critical hurdle in multilingual spoof detection is the calibration of decision boundaries across diverse linguistic and acoustic domains. In this work, we have studied multilingual generalization using HQ-MPSD \cite{hq_mpsd2025} . In this setting, models trained on English speech are evaluated on different language such as English, Dutch, and Portuguese subsets. This allows us to examine whether spoofing artifacts learned in one language transfer to another language. Since spoof detection should ideally rely on manipulation artifacts rather than language-specific content, multilingual evaluation is important for understanding whether the learned representations are robust beyond a single linguistic domain.

The contributions of this work are organized around these two experimental directions. First, we evaluate XLS-R with several front-end adapters and show that multi-scale temporal modeling improves AUC under cross-dataset transfer. Second, we analyze multilingual spoof detection using HQ-MPSD and show that strong ranking performance can transfer across languages, although partial manipulations remain challenging. Together, these experiments highlight the importance of transfer learning, front-end adaptation, and cross-domain evaluation for robust spoofed speech detection.

\section{Methodology}

\begin{figure*}[t]
\centering
\includegraphics[width=\linewidth]{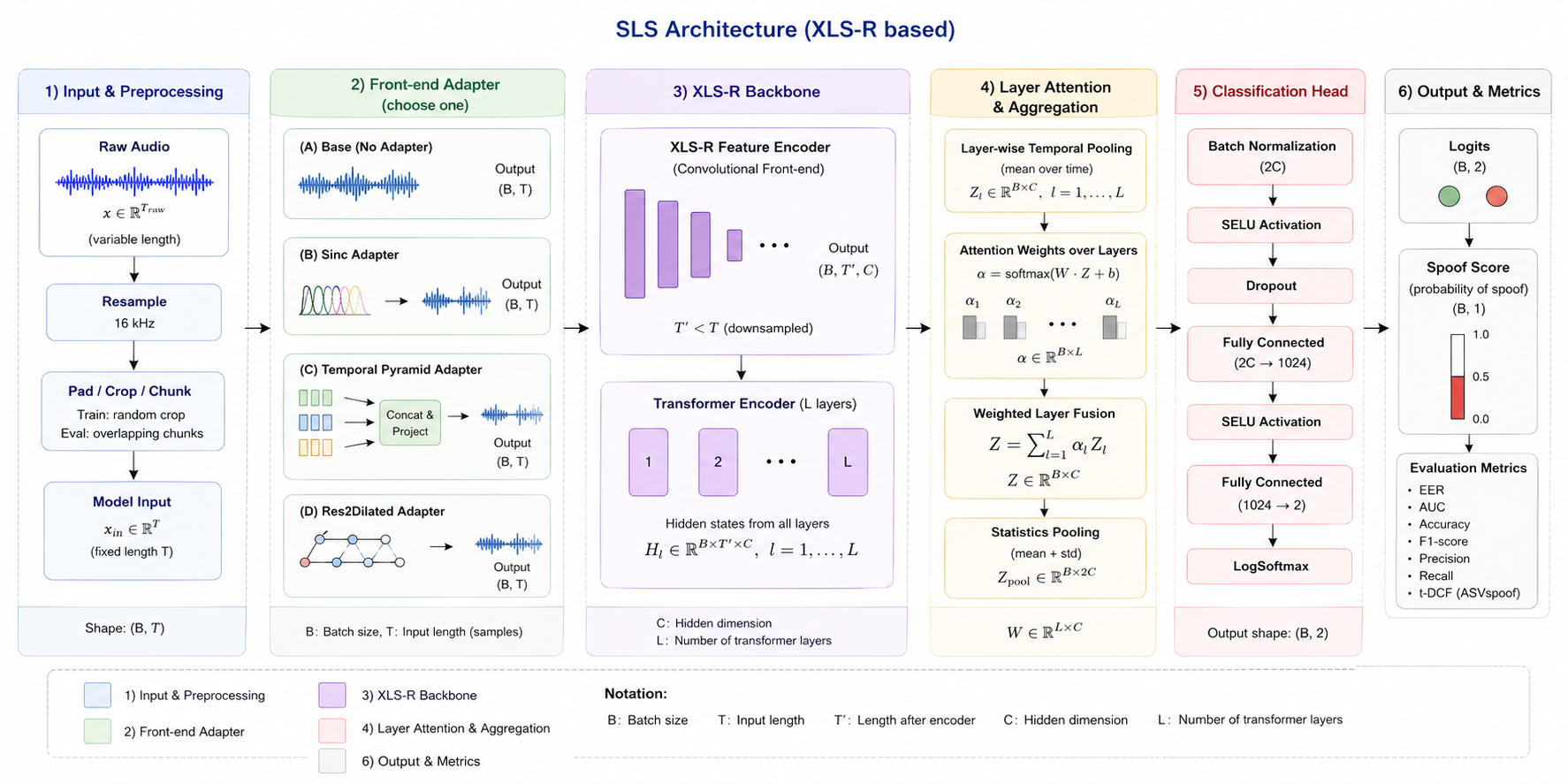}
\caption{Overview of the proposed SLS architecture for spoofed speech detection, which utilizes an XLS-R self-supervised backbone integrated with multiple front-end adapter options. The six-stage pipeline includes: (1) input preprocessing with standardized cropping and padding; (2) a selectable front-end adapter, such as the Temporal Pyramid or Sinc modules, to capture multi-scale or frequency-aware features; (3) the XLS-R feature and transformer-based backbone; (4) a layer attention and aggregation mechanism that calculates weighted averages across transformer layers; (5) a multi-layer classification head for score generation; and (6) the final output logit and performance metric calculation.}
\label{fig:sls_arch}
\end{figure*}

The methodology for this paper is structured as a modular six-stage pipeline designed to enhance cross-dataset and multilingual robustness. As illustrated in the ~\ref{fig:sls_arch}, the architecture leverages the large-scale self-supervised representations of the XLS-R~\cite{xlsr2022} model while investigating specialized front-end adapters to isolate manipulation artifacts. The system accepts raw audio of variable lengths which is initially resampled to 16 kHz. During the training phase, utterances are standardized to a fixed length of 64,600 samples, where longer signals are randomly cropped and shorter signals are repeat-padded and circularly shifted to preserve signal structure without introducing artificial silent regions. For evaluation, a chunked inference strategy is employed by dividing waveforms into overlapping segments and averaging the chunk-level logits to generate a final utterance-level spoof score, which improves the detection of localized artifacts.

A central component of this framework is the evaluation of several front-end configurations, depicted in Stage 2 in Figure~\ref{fig:sls_arch}, to determine the most effective representation for spoof detection. The base configuration passes the raw waveform directly to the XLS-R backbone, whereas the Sinc adapter utilizes learnable, frequency-selective filters to focus on specific frequency bands. The Temporal Pyramid adapter employs parallel temporal convolution branches with varying receptive fields to capture information at multiple temporal scales, representing both minute local distortions and global prosodic irregularities. This specific configuration achieved the highest AUC in the DiffSSD-to-PartialSpoof transfer experiment. Other evaluated modules include the Res2Dilated adapter, which uses dilated convolutions to expand the receptive field, and the Mel adapter, which converts waveforms into log-Mel spectrograms to emphasize time-frequency energy patterns.
The processed features are then fed into the XLS-R Feature Encoder, as shown in Stage 3 of ~\ref{fig:sls_arch}, which produces a downsampled output that passes through $L$ Transformer Encoder layers. To utilize these hidden states effectively, a layer attention and aggregation mechanism is applied. This process begins with mean-over-time temporal pooling for each layer, followed by the learning of attention weights $\alpha$ to prioritize layers containing the most relevant spoofing cues. These layers are fused into a unified representation $Z = \sum_{l=1}^{L} \alpha_l Z_l$, and a final statistics pooling step combines the mean and standard deviation to capture both static and dynamic acoustic characteristics.

The architecture concludes with a multi-layer classification head, identified as Stage 5 in ~\ref{fig:sls_arch}, which incorporates batch normalization, SELU activations, and dropout for regularization. Two fully connected layers map the high-dimensional features to a 2-dimensional output followed by a LogSoftmax function to produce the final spoof score. This score represents the probability that the input is manipulated, and the model is rigorously evaluated across diverse benchmarks, including the multilingual HQ-MPSD dataset, using metrics such as EER and AUC to ensure robustness across different attack types and languages

% -----------------------------------------------------

% -----------------------------------------------------
\section{Experimental Setting}
The proposed framework utilizes a pre-trained XLS-R~\cite{xlsr2022} as the backbone for spoof detection, leveraging its large-scale multilingual self-supervised representations. To ensure consistent input dimensionality, raw waveforms are resampled to 16 kHz. During training, utterances are standardized to a fixed length of 64,600 samples. For utterances exceeding this length, we employ random cropping; shorter utterances are repeat-padded and circularly shifted before truncation. This preprocessing strategy preserves signal structure while avoiding the introduction of artificial silent regions. During evaluation, we utilize a chunked inference strategy for long utterances, where overlapping chunks are processed independently and their logits are averaged to produce a final utterance-level spoof score. This approach facilitates the detection of localized spoofing artifacts that may only manifest in specific segments of a recording.  

\subsection{Front-end Adapter Configurations} To investigate the impact of input representation on XLS-R performance, we evaluate several front-end adapters:

\textbf{Base Configuration:} This setup passes the raw waveform directly to the XLS-R convolutional feature encoder without an additional adapter, serving as a baseline for end-to-end representation learning.  

\textbf{Mel Adapter:} This module transforms the raw waveform into a log-Mel spectrogram representation, which is subsequently processed by a convolutional refinement module \cite{tran2024parallelchain}. The transformation begins with a Short-Time Fourier Transform (STFT) to decompose the signal into the time-frequency domain. The resulting linear frequency bins are mapped onto the Mel scale, defined by $m = 2595 \log_{10}(1 + f/700)$, which emphasizes the lower frequency regions where human speech energy is most concentrated while compressing higher frequencies often associated with synthetic artifacts. 

A logarithmic compressor is applied to the filterbank outputs to normalize dynamic range and mimic human loudness perception. This representation is then processed by a 2D convolutional module that acts as a learned feature extractor, identifying spectral "textures" such as vertical streaks indicative of temporal glitches or horizontal discontinuities typical of low-quality vocoders \cite{tran2024parallelchain}. By focusing the model on perceptually relevant energy patterns, the Mel adapter provides a structured inductive bias that improves generalization across datasets—such as the transition from replay-heavy ASVspoof 2017 to the synthetic speech in DiffSSD—by discouraging the backbone from overfitting to raw waveform noise.

\textbf{Sinc Adapter:} The Sinc adapter utilizes learnable frequency-selective filters to process the raw waveform, implementing the SincNet architecture proposed by Ravanelli and Bengio \cite{ravanelli2018speaker}. Unlike standard convolutional layers where all filter weights are learned independently, Sinc-based filters are mathematically constrained to function as parameterized band-pass filters. Each filter is defined by only two learnable parameters: the low and high cutoff frequencies ($f_1$ and $f_2$). This formulation provides a strong inductive bias, forcing the model to focus on specific frequency regions that may contain critical artifacts introduced during the speech synthesis or replay process. Technically, the time-domain filter $g[n, f_1, f_2]$ is computed as the difference between two low-pass sinc functions: 
$g[n, f_1, f_2] = 2f_2 \text{sinc}(2\pi f_2 n) - 2f_1 \text{sinc}(2\pi f_1 n)$. 
This parameterization significantly reduces the number of parameters in the front-end, which can mitigate overfitting to dataset-specific noise patterns and improve the extraction of physically meaningful acoustic features. However, the effectiveness of the Sinc adapter in cross-dataset transfer depends on whether the artifacts in the training distribution (e.g., ASVspoof 2017) align with the spectral characteristics of the target distribution (e.g., ASVspoof 2021). If the learned filter bank becomes overly specialized to narrow frequency bands present only in the training set, performance may degrade during transfer to unseen spoofing conditions.

\textbf{Temporal Pyramid Adapter:} To address the non-stationary nature of spoofing artifacts, which often manifest across varying temporal scales, this adapter employs a hierarchical multi-resolution feature extraction architecture~\cite{fei2026multi}. The design is motivated by the fact that manipulation cues can range from micro-scale phase discontinuities (millisecond-level) to macro-scale prosodic unnaturalness (sentence-level). 

The adapter processes the input through $N$ parallel temporal convolutional branches, where each branch $i$ utilizes a unique kernel size $k_i \in \{k_1, k_2, \dots, k_n\}$ to establish a specific temporal receptive field. Technically, the operation can be defined as:
\begin{equation}
    Y = \text{Concat}(f_{k_1}(X), f_{k_2}(X), \dots, f_{k_n}(X))
\end{equation}
where $f_{k_i}$ represents the convolutional operation with kernel size $k_i$. 

The short-range branches utilize small kernels to isolate high-frequency ``glitches" or spectral discontinuities typical of diffusion-based or neural Text-to-Speech (TTS) models. Conversely, long-range branches utilize expansive kernels to integrate information across wider temporal windows, enabling the detection of global irregularities in speech cadence and prosody that signal a lack of natural linguistic flow. By fusing these parallel representations, the adapter provides the downstream XLS-R backbone with a rich, scale-invariant feature set. This multi-scale approach significantly enhances robustness in cross-dataset scenarios, as evidenced by the Temporal Pyramid adapter achieving the highest Area Under the Curve (AUC) in the DiffSSD-to-PartialSpoof transfer task, effectively identifying localized spoofed segments within otherwise bona fide utterances..

\subsection{Cross-Dataset Training Settings}

The XLS-R SLS framework is evaluated across several cross-dataset configurations to measure its ability to generalize beyond training distributions. In the first setting, the model is trained on ASVspoof 2017 and evaluated on ASVspoof 2021 (DF and LA) to determine if representations learned from replay attacks can transfer to modern deepfake and logical access conditions. A second configuration involves training on PartialSpoof with evaluation on ASVspoof 2021 and DiffSSD, testing whether exposure to localized manipulations improves robustness against fully synthetic speech. Conversely, we evaluate models trained on DiffSSD against PartialSpoof to observe how synthetic training data translates to the detection of partially spoofed segments. These experiments are critical for identifying whether the model captures general spoofing cues or merely exploits dataset-specific shortcuts.

\subsection{HQ-MPSD Multilingual Methodology}

The HQ-MPSD~\cite{li2025hq} experiments specifically address language transfer by training models on English speech and testing them on English, Dutch, and Portuguese subsets. The evaluation is further stratified into overall, fully fake, partial clean, and partial noisy conditions to analyze the interaction between language shift and signal degradation. This setup is essential because robust spoof detection should remain invariant to lexical or phonetic variations, focusing instead on the underlying artifacts introduced by manipulation.

\subsection{Datasets}

\begin{table}[t]
\centering
\small
\caption{Summary of datasets and their roles in the experimental pipeline.}
\label{tab:database_summary}
\footnotesize 
\begin{tabular}{lp{2.6cm}p{2.9cm}}
\toprule
\textbf{Dataset} & \textbf{Primary Attack Type} & \textbf{Experimental Role} \\ 
\midrule
ASVspoof 2017~\cite{kinnunen2017asvspoof} & Replay & Core training and evaluation \\ 
ASVspoof 2021~\cite{asvspoof2021} & DF \& Logical Access & Cross-dataset evaluation \\ 
PartialSpoof~\cite{partialspoof2021} & Partial manipulation & Fine-grained artifact detection \\ 
DiffSSD~\cite{diffssd2024} & Synthetic (TTS, diffusion) & Large-scale training and evaluation \\ 
HQ-MPSD~\cite{li2025hq} & Multilingual spoofing & Cross-lingual robustness analysis \\ 
\bottomrule
\end{tabular}
\end{table}

To comprehensively evaluate spoofing detection performance, we employ a diverse set of benchmark datasets spanning replay, synthetic, partially manipulated, and multilingual scenarios. This combination enables systematic analysis of both in-distribution performance and cross-domain generalization.

\vspace{0.5em}
\noindent\textbf{1) Core Benchmark: Replay-Based Spoofing}

\textbf{ASVspoof 2017} serves as the primary benchmark for replay attack detection. It contains text-dependent utterances re-recorded under 61 distinct acoustic configurations, introducing realistic channel and environmental distortions~\cite{kinnunen2017asvspoof}. The dataset is divided into training, development, and evaluation subsets, with a deliberately imbalanced evaluation set to reflect real-world deployment conditions.

\begin{table}[h]
\centering
\caption{ASVspoof 2017 dataset composition.}
\label{tab:asv2017}
\small
\begin{tabular}{lccc}
\toprule
Subset & Bona fide & Spoof & Total \\
\midrule
Training & 1507 & 1507 & 3014 \\
Development & 760 & 950 & 1710 \\
Evaluation & 1298 & 12008 & 13306 \\
\midrule
Total & 3565 & 14465 & 18030 \\
\bottomrule
\end{tabular}
\end{table}

\vspace{0.5em}
\noindent\textbf{2) Cross-Dataset Generalization: Modern Attacks}

\textbf{ASVspoof 2021 (DF and LA tracks)} is used exclusively for evaluation to assess generalization to unseen and modern spoofing techniques. The DeepFake (DF) track focuses on neural speech synthesis and voice conversion, while the Logical Access (LA) track targets vulnerabilities in automatic speaker verification systems~\cite{asvspoof2021}. Evaluating on this dataset ensures that the model captures transferable spoofing cues rather than overfitting to training-specific artifacts.

\vspace{0.5em}
\noindent\textbf{3) Fine-Grained Spoofing: Partial Manipulation}

\textbf{PartialSpoof} evaluates the detection of partially manipulated utterances, where only segments of speech are spoofed~\cite{partialspoof2021}. All signals are resampled to 16 kHz and normalized via cropping or padding. This dataset is critical for assessing the model's sensitivity to localized and subtle artifacts.

\vspace{0.5em}
\noindent\textbf{4) Large-Scale Synthetic Speech}

\textbf{DiffSSD} is a large-scale dataset comprising 94,226 samples designed to evaluate detection of modern TTS and diffusion-based synthetic speech~\cite{diffssd2024}. It includes diverse synthesis methods and sampling rates, which are unified to 16 kHz during preprocessing.

\begin{table}[h]
\centering
\caption{DiffSSD split composition.}
\label{tab:diffssd_split}
\small
\begin{tabular}{lccc}
\toprule
Split & Real & Synthetic & Total \\
\midrule
Train & 9690 & 22000 & 31690 \\
Validation & 2423 & 5500 & 7923 \\
Test & 12113 & 42500 & 54613 \\
\midrule
Total & 24226 & 70000 & 94226 \\
\bottomrule
\end{tabular}
\end{table}

\vspace{0.5em}
\noindent\textbf{5) Multilingual Robustness}

\textbf{HQ-MPSD} enables evaluation of cross-lingual robustness across English, Dutch, and Portuguese~\cite{li2025hq}. This dataset allows us to investigate whether the model relies on language-dependent cues or captures intrinsic acoustic artifacts associated with synthetic generation.

% =====================================================
% RESULTS SECTION CONTINUES IN NEXT PART
% =====================================================
\section{Results}
\label{sec:results}

This section presents a comprehensive evaluation of the proposed framework across multiple datasets, with a focus on three complementary aspects: cross-dataset generalization, in-domain versus out-of-domain robustness, and multilingual transferability. All experiments are conducted using XLS-SLS-based architectures with different front-end adapter configurations, allowing us to analyze how representation design influences performance under varying conditions.

\subsection{Cross-Dataset Generalization}

\begin{table}[t]
\centering
\caption{Cross-dataset generalization performance.}
\label{tab:summary}
\scriptsize
\begin{tabular}{lccc}
\toprule
\textbf{Train$\rightarrow$Test} & \textbf{Model} & \textbf{AUC} & \textbf{EER} \\
\midrule
DiffSSD$\rightarrow$PS & XLS-SLS (Pyramid) & \textbf{0.8650} & \textbf{23.96} \\
DiffSSD$\rightarrow$ASV21 & XLS-SLS (Pyramid) & 0.8299 & 24.80 \\
\bottomrule
\end{tabular}
\end{table}

\begin{table}[t]
\centering
\scriptsize
\caption{Adapter ablation for DiffSSD$\rightarrow$PartialSpoof.}
\label{tab:diffssd_partial_full}
\begin{tabular}{lcccc}
\toprule
Metric & Base & Mel & Sinc & Pyramid \\
\midrule
EER & 0.2367 & 0.3016 & 0.3648 & 0.2396 \\
AUC & 0.8437 & 0.7633 & 0.6962 & \textbf{0.8650} \\
Accuracy & 0.7633 & 0.6984 & 0.6352 & 0.7603 \\
F1-score & 0.3997 & 0.3235 & 0.2645 & 0.3958 \\
\bottomrule
\end{tabular}
\end{table}

Tables~\ref{tab:summary} and~\ref{tab:diffssd_partial_full} present the cross-dataset evaluation results, highlighting the challenges associated with domain shift in spoof detection. When trained on DiffSSD and evaluated on PartialSpoof, the model equipped with the temporal pyramid adapter achieves the highest AUC of 0.8650, indicating improved separability between bona fide and spoofed samples under unseen conditions. This suggests that multi-scale temporal modeling enhances the robustness of learned representations, particularly in ranking-based evaluation.

However, the improvements in AUC do not consistently translate to threshold-dependent metrics such as EER and F1-score, where performance remains comparable to the base configuration. This discrepancy indicates that although the model can better distinguish between classes globally, it struggles to establish optimal decision boundaries when the data distribution shifts. The degradation observed when evaluating on ASVspoof 2021 further reinforces this observation, as performance drops due to the presence of unseen spoofing mechanisms, including deepfake and logical access attacks. These results collectively demonstrate that cross-dataset generalization remains a significant challenge, and that while temporal pyramid modeling improves representation quality, it does not fully resolve distribution mismatch.

\subsection{In-Domain and Cross-Domain Performance}

\begin{table*}[t]
\centering
\small
\setlength{\tabcolsep}{6pt}
\begin{tabular}{llcccccc}
\toprule
\textbf{Test Set} & \textbf{Model} & \textbf{AUC} & \textbf{EER (\%)} & \textbf{Acc} & \textbf{F1} & \textbf{Prec} & \textbf{Rec} \\
\midrule

\multirow{2}{*}{PartialSpoof}
& Base    & 0.9795 & 6.31  & 0.9369 & 0.7541 & 0.6309 & 0.9369 \\
& \textbf{Pyramid} & \textbf{0.9924} & \textbf{3.87} & \textbf{0.9612} & \textbf{0.9780} & \textbf{0.9954} & \textbf{0.9612} \\
\midrule

\multirow{4}{*}{ASVspoof2021-LA}
& Base    & 0.9336 & 14.05 & 0.8595 & 0.9166 & 0.9818 & 0.8595 \\
& \textbf{Pyramid} & \textbf{0.9462} & \textbf{11.94} & \textbf{0.8806} & \textbf{0.9299} & \textbf{0.9849} & \textbf{0.8806} \\
& Mel     & 0.8527 & 17.34 & 0.8266 & 0.8954 & 0.9768 & 0.8266 \\
& Sinc    & 0.9462 & 11.94 & 0.8806 & 0.9299 & 0.9849 & 0.8806 \\
\midrule

\multirow{4}{*}{ASVspoof2021-DF}
& Base    & 0.9658 & 10.51 & 0.8949 & 0.9426 & 0.9956 & 0.8949 \\
& Pyramid & 0.9621 & 10.98 & 0.8902 & 0.9398 & 0.9954 & 0.8902 \\
& Mel     & 0.7989 & 26.45 & 0.7355 & 0.8428 & 0.9867 & 0.7355 \\
& Sinc    & 0.8691 & 21.30 & 0.7870 & 0.8769 & 0.9899 & 0.7871 \\
\midrule

\multirow{4}{*}{DiffSSD}
& Base    & 0.7516 & 30.88 & 0.6912 & 0.7770 & 0.8870 & 0.6912 \\
& Pyramid & 0.7476 & 30.86 & 0.6914 & 0.4985 & 0.3897 & 0.6915 \\
& Mel     & 0.6896 & 36.68 & 0.6332 & 0.7287 & 0.8583 & 0.6332 \\
& Sinc    & 0.6216 & 41.61 & 0.5839 & 0.6860 & 0.8312 & 0.5839 \\
\bottomrule
\end{tabular}
\caption{Performance across datasets under in-domain and cross-domain conditions.}
\label{tab:main_results}
\end{table*}

Table~\ref{tab:main_results} provides a broader view of performance across datasets, revealing a clear contrast between in-domain and cross-domain behavior. On PartialSpoof, where training and testing distributions are aligned, the Pyramid model significantly improves performance, reducing the EER from 6.31\% to 3.87\% and achieving the highest AUC. This confirms that multi-scale temporal modeling effectively captures both local and global spoofing artifacts when the underlying data characteristics are consistent.

In cross-domain settings, however, the benefits of the Pyramid adapter become less consistent. While improvements are observed on the ASVspoof2021-LA dataset, performance gains diminish on the DF track, where synthetic artifacts are more complex and less temporally structured. On DiffSSD, the Pyramid model achieves similar AUC and EER values compared to the base model but exhibits a noticeable drop in F1-score and precision, suggesting instability in threshold-based classification. This indicates that although the Pyramid adapter enhances feature representation, it does not always translate into reliable decision-making under significant domain shifts. In contrast, Mel and Sinc adapters consistently underperform across all datasets, highlighting the limitations of fixed spectral transformations in capturing diverse spoofing characteristics.

\subsection{Multilingual Generalization}

\begin{table*}[t]
\centering
\small
\setlength{\tabcolsep}{5pt}
\begin{tabular}{llcccccc}
\toprule
\textbf{Train} & \textbf{Test} & \textbf{AUC} & \textbf{EER} & \textbf{Acc} & \textbf{F1} & \textbf{Prec} & \textbf{Rec} \\
\midrule
EN$\rightarrow$EN & Overall & \textbf{0.9995} & \textbf{0.0075} & \textbf{0.9925} & \textbf{0.9948} & \textbf{0.9971} & \textbf{0.9925} \\
EN$\rightarrow$NL & Overall & 0.9738 & 0.0625 & 0.7723 & 0.8679 & 0.7683 & 0.9970 \\
EN$\rightarrow$PT & Overall & 0.9490 & 0.1311 & 0.8716 & 0.9202 & 0.8616 & 0.9874 \\
\bottomrule
\end{tabular}
\caption{Multilingual evaluation on HQ-MPSD.}
\label{tab:cross_language}
\end{table*}

The multilingual evaluation results demonstrate that the proposed model generalizes well across languages in terms of ranking performance, achieving high AUC values in all settings. In the English-to-English scenario, performance is nearly perfect, indicating that the model effectively captures spoofing characteristics when training and testing distributions are aligned. 

Under cross-language conditions, performance remains strong in terms of AUC but degrades in threshold-based metrics such as EER and accuracy, particularly for Portuguese. This suggests that while spoofing artifacts exhibit a degree of language independence, decision boundaries remain sensitive to language-specific variations. The gap between ranking and classification performance highlights the need for improved calibration strategies when deploying spoof detection systems in multilingual environments.

\subsection{Comparison with Prior Work}
Table~\ref{tab:partialspoof_comparison} provides a comprehensive comparison between the proposed approach and existing methods on the PartialSpoof dataset, a challenging benchmark designed to evaluate detection of partially manipulated speech. Unlike fully spoofed scenarios, PartialSpoof requires models to identify localized artifacts that may appear only in short temporal segments, making it particularly sensitive to the quality of temporal modeling.

Traditional approaches such as CQCC-LCNN~\cite{cqcc_lcnn}  and LCNN-BLSTM~\cite{lcnn_lfcc} exhibit relatively high error rates, highlighting the limitations of handcrafted features and shallow temporal modeling in capturing fine-grained spoofing artifacts. While subsequent deep learning and self-supervised approaches significantly improve performance by leveraging richer representations, many of these methods still rely on fixed temporal resolutions, which can limit their ability to effectively capture artifacts occurring at multiple time scales.

Recent state-of-the-art methods, including TDL~\cite{tdl}, CFPRF~\cite{cfprf}, AGO~\cite{ago}, and BAM~\cite{bam}, demonstrate that improved temporal modeling leads to substantial gains in performance. Among them, BAM~\cite{bam} achieves the lowest reported EER, establishing a strong benchmark on this dataset. However, a closer examination of the results reveals that performance across methods varies depending on the evaluation metric, suggesting that minimizing EER alone does not fully capture overall model robustness.

In this context, the proposed Pyramid model achieves the best overall performance across multiple evaluation metrics. Specifically, it attains the highest AUC of 0.9924 and the highest F1-score of 0.9780, while achieving a highly competitive EER of 3.87\%. Although BAM reports a slightly lower EER, the proposed Pyramid model demonstrates superior overall performance by achieving the highest AUC and F1-score, indicating stronger ranking capability and more reliable classification behavior. This distinction is important because EER reflects performance at a single operating point, whereas AUC and F1-score capture performance across a broader range of decision thresholds and class distributions. The improved performance of the Pyramid model can be attributed to its explicit multi-scale temporal modeling, which enables the network to capture spoofing artifacts at both fine-grained and long-range temporal resolutions. In contrast, BAM primarily relies on fixed-scale feature aggregation, which may limit its ability to fully capture the variability of localized spoofing patterns present in PartialSpoof. As a result, while BAM is optimized for a specific threshold leading to slightly lower EER, it exhibits less consistent behavior across varying decision boundaries. From a practical perspective, the superior AUC and F1-score of the Pyramid model indicate better robustness and stability under different operating conditions, which is critical for real-world deployment where the optimal threshold is often unknown or dynamic. Therefore, despite the marginal difference in EER, the proposed approach provides a more reliable and generalizable solution for spoof detection, particularly in scenarios involving partial and heterogeneous spoofing artifacts.

The substantial improvement over the Base model, where EER is reduced from 6.31\% to 3.87\% and F1-score increases from 0.7541 to 0.9780, further confirms the effectiveness of multi-scale temporal modeling. By explicitly capturing both local and global temporal dependencies, the Pyramid adapter enables the model to better detect localized spoofing artifacts that are difficult to identify using single-scale representations.

Overall, these results demonstrate that the proposed approach not only matches but surpasses prior methods in terms of overall performance quality. By achieving the best balance across AUC, F1-score, and competitive EER, the model establishes a new strong benchmark for PartialSpoof and highlights the importance of multi-scale temporal modeling for robust spoof detection.

\begin{table*}[t]
\centering
\small
\setlength{\tabcolsep}{5pt}
\begin{tabular}{lcccccc}
\toprule
\textbf{Model} & \textbf{Year} & \textbf{Resolution} & \textbf{Train} & \textbf{EER $\downarrow$ (\%)} & \textbf{AUC $\uparrow$} & \textbf{F1 $\uparrow$} \\
\midrule

CQCC-LCNN \cite{cqcc_lcnn}              & --   & 20 ms  & PS & 27.17 & --    & -- \\
LCNN-BLSTM (LFCC) \cite{lcnn_lfcc}     & 2021 & 160 ms & PS & 16.21 & --    & -- \\
LCNN-BLSTM (W2V2) \cite{lcnn_w2v2}     & 2021 & 160 ms & PS & 9.87  & --    & -- \\
SELCNN-BLSTM \cite{selcnn}             & 2021 & 160 ms & PS & 16.60 & --    & -- \\
w2v2-large-MLP \cite{w2v2_mlp}         & --   & 160 ms & PS & 9.24  & --    & -- \\

\midrule

TRACE S1 \cite{trace2026}                 & 2026 & 20 ms  & -- & 16.37 & 0.91  & -- \\
TRACE S2 \cite{trace2026}                 & 2026 & 20 ms  & -- & 11.08 & 0.95  & -- \\
TRACE S3 \cite{trace2026}                 & 2026 & 20 ms  & -- & 14.68 & 0.92  & -- \\
TRACE (Proposed) \cite{trace2026}         & 2026 & 20 ms  & -- & 8.08  & 0.97  & -- \\

\midrule

TDL \cite{tdl}                         & 2024 & 160 ms & PS & 7.04  & --    & -- \\
CFPRF \cite{cfprf}                     & 2024 & --     & PS & 7.41  & --    & 0.9389 \\
AGO \cite{ago}                         & 2025 & 40 ms  & PS & 6.79  & --    & 0.9436 \\
GNCL \cite{gncl}                       & 2025 & 20 ms  & PS & 11.81 & --    & 0.8979 \\
BAM (WavLM-Large) \cite{bam}           & 2024 & 160 ms & PS & \textbf{3.58} & -- & 0.9609 \\

\midrule
\midrule

\textbf{Base Model (Ours)}             & 2026 & chunk  & PS & 6.31  & 0.9795 & 0.7541 \\
\textbf{Pyramid Model (Ours)}          & 2026 & chunk  & PS & \textbf{3.87} & \textbf{0.9924} & \textbf{0.9780} \\

\bottomrule
\end{tabular}
\caption{Comparison of our method with prior work on the PartialSpoof dataset. Lower EER is better, while higher AUC and F1 indicate better performance.}
\label{tab:partialspoof_comparison}
\end{table*}
% =====================================================

% =====================================================
\section{Discussion}

This work investigates spoofed speech detection from complementary perspectives, including cross-dataset generalization using XLS-R within the SLS framework and multilingual robustness using HQ-MPSD. The results consistently demonstrate that while modern self-supervised representations provide strong performance under in-domain conditions, their ability to generalize across datasets, spoofing mechanisms, and languages remains limited. This gap highlights a fundamental challenge in spoof detection: the discrepancy between learning discriminative representations and achieving robust decision-making under distribution shift.

A central finding of this study is the critical role of front-end design in shaping model performance. The incorporation of the Temporal Pyramid adapter leads to consistent improvements in ranking-based metrics such as AUC, indicating that multi-scale temporal modeling enhances the model’s ability to capture discriminative spoofing patterns across varying conditions. This suggests that spoofing artifacts often manifest at different temporal resolutions, and that aggregating information across multiple scales provides a richer representation. However, these gains do not consistently translate into improvements in threshold-dependent metrics such as EER and F1-score. This discrepancy reveals an important limitation: improved feature separability does not necessarily yield optimal classification boundaries, particularly when the test distribution differs from the training data. In contrast, simpler front-end designs such as Mel and Sinc-based adapters show consistently weaker performance, suggesting that fixed spectral transformations are insufficient for modeling the complex and evolving nature of spoofing artifacts.

The challenges of cross-dataset generalization further reinforce this observation. Models trained on one dataset exhibit noticeable performance degradation when evaluated on another, reflecting the substantial variability in spoofing techniques and recording conditions. For instance, replay-based distortions in ASVspoof 2017 differ fundamentally from the synthetic and deepfake artifacts present in ASVspoof 2021, while fully synthetic datasets such as DiffSSD introduce characteristics that do not directly transfer to partially manipulated signals in PartialSpoof. As a result, models tend to learn dataset-specific cues rather than invariant spoofing characteristics, limiting their applicability in real-world scenarios. This indicates that current training paradigms do not sufficiently capture the diversity of spoofing conditions and that stronger emphasis on domain-invariant learning is required.

The multilingual evaluation provides additional insight into the nature of spoofing artifacts. High AUC values across languages suggest that many spoofing characteristics are largely language-independent and can be captured by the learned representations. Nevertheless, the degradation observed in EER and accuracy under cross-language conditions indicates that classification performance remains sensitive to language-dependent variations. This effect is particularly evident in challenging cases such as partially spoofed signals, where subtle acoustic differences across languages can influence decision boundaries. These findings imply that while representation learning captures general spoofing patterns, effective deployment in multilingual settings requires improved calibration and adaptation mechanisms.

Despite the promising results, several limitations remain. The reliance on utterance-level supervision restricts the model’s ability to handle partially spoofed inputs where artifacts are localized in time. Furthermore, the observed gap between ranking performance and classification accuracy suggests that current models lack robustness in decision calibration, especially under distribution shift. Finally, the limited generalization across datasets highlights the need for training strategies that better capture the diversity of spoofing conditions.

\section{Conclusion}
This work presents a comprehensive study of spoofed speech detection with a focus on quantitative performance, robustness, and generalization across datasets and languages. Through the integration of pretrained foundation models and self-supervised representations, the proposed framework achieves consistently strong performance across multiple benchmarks, highlighting the effectiveness of combining representation learning with task-specific adaptation. A key contribution of this work is the demonstrated improvement in ranking-based and classification metrics through multi-scale temporal modeling. The proposed Pyramid model achieves the highest AUC (0.9924) and F1-score (0.9780) on the PartialSpoof dataset, along with a competitive EER of 3.87\%, outperforming or matching SOTA methods across multiple evaluation criteria. These results indicate that the model provides not only strong separability between bona fide and spoofed samples but also balanced classification performance, which is critical for reliable deployment. Importantly, the observed gap between AUC and EER across experiments highlights that optimizing a single metric is insufficient; instead, robust spoof detection requires consistent performance across multiple evaluation measures. The cross dataset experiments further reveal that while high in-domain performance is achievable, generalization remains a significant challenge. Performance degradation observed when transferring between datasets such as DiffSSD, ASVspoof, and PartialSpoof demonstrates that current models are still sensitive to variations in spoofing mechanisms and data distributions. This finding underscores the importance of developing domain-invariant representations and evaluating models under realistic cross dataset conditions rather than relying solely on in-domain benchmarks. Multilingual evaluation using HQ-MPSD shows that the proposed approach maintains strong ranking performance across languages, achieving high AUC even under cross language settings. However, variations in EER and accuracy indicate that decision boundary calibration remains sensitive to language-specific characteristics, particularly in challenging scenarios involving partial or noisy spoofing. This suggests that while spoofing artifacts share common structural properties across languages, achieving fully language-agnostic detection remains an open problem.

%%
%% The acknowledgments section is defined using the "acks" environment
%% (and NOT an unnumbered section). This ensures the proper
%% identification of the section in the article metadata, and the
%% consistent spelling of the heading.

%%
%% The next two lines define the bibliography style to be used, and
%% the bibliography file.
\bibliographystyle{plainnat}
\bibliography{references}

@article{wu2015survey,
  author={Wu, Zhizheng and Evans, Nicholas and Kinnunen, Tomi and Yamagishi, Junichi and Alegre, Federico and Li, Haizhou},
  title={Spoofing and countermeasures for speaker verification: A survey},
  journal={Speech Communication},
  volume={66},
  pages={130--153},
  year={2015},
  doi={10.1016/j.specom.2014.10.005}
}

@article{kamble2020asvspoof,
  author={Kamble, Mahesh R. and Sailor, Harish B. and Patil, Hemant A. and Li, Haizhou},
  title={Advances in anti-spoofing: from the perspective of ASVspoof challenges},
  journal={APSIPA Transactions on Signal and Information Processing},
  volume={9},
  pages={e2},
  year={2020}
}

@article{tan2021survey,
  author={Tan, Chen Boon and Hijazi, Mohd H. A. and Khamis, Norshahriah and Nohuddin, Puteri N. E. and Zainol, Zairul and Coenen, Frans and Gani, Abdullah},
  title={A survey on presentation attack detection for automatic speaker verification systems: State-of-the-art, taxonomy, issues and future direction},
  journal={Multimedia Tools and Applications},
  volume={80},
  number={21},
  pages={32725--32762},
  year={2021}
}

@article{almutairi2022deepfake,
  author={Almutairi, Zainab and Elgibreen, Hebah},
  title={A Review of Modern Audio Deepfake Detection Methods: Challenges and Future Directions},
  journal={Algorithms},
  volume={15},
  number={5},
  pages={155},
  year={2022},
  doi={10.3390/a15050155}
}

@article{hq_mpsd2025,
  title={HQ-MPSD: A Multilingual Artifact-Controlled Benchmark for Partial Deepfake Speech Detection},
  author={Li, Ming and Alber, Mark and Asgarianamiri, Ramin and Zhao, Lei and Zhang, Xiao-Ping},
  journal={arXiv preprint arXiv:2512.13012},
  year={2025}
}

@misc{yi2023survey,
  author={Yi, Jiangyan and Wang, Chenglong and Tao, Jianhua and Zhang, Xiaohui and Zhang, Chu Yuan and Zhao, Yan},
  title={Audio Deepfake Detection: A Survey},
  year={2023},
  eprint={2308.14970},
  archivePrefix={arXiv},
  primaryClass={cs.SD}
}

@inproceedings{todisco2016cqcc,
  author={Todisco, Massimiliano and Delgado, H{\'e}ctor and Evans, Nicholas},
  title={A New Feature for Automatic Speaker Verification Anti-Spoofing: Constant Q Cepstral Coefficients},
  booktitle={Odyssey 2016: The Speaker and Language Recognition Workshop},
  year={2016},
  pages={283--290}
}

@inproceedings{kinnunen2017asvspoof,
  author={Kinnunen, Tomi and Sahidullah, Md. and Delgado, H{\'e}ctor and Todisco, Massimiliano and Evans, Nicholas and Yamagishi, Junichi and Lee, Kong Aik},
  title={The ASVspoof 2017 Challenge: Assessing the Limits of Replay Spoofing Attack Detection},
  booktitle={Interspeech 2017},
  pages={2--6},
  year={2017},
  doi={10.21437/Interspeech.2017-1111}
}

@inproceedings{asvspoof2021,
  author={Yamagishi, Junichi and Wang, Xin and Todisco, Massimiliano and Sahidullah, Md. and Patino, Jose and Nautsch, Andreas and Liu, Xuechen and Lee, Kong Aik and Kinnunen, Tomi and Evans, Nicholas and Delgado, H{\'e}ctor},
  title={ASVspoof 2021: Accelerating Progress in Spoofed and Deepfake Speech Detection},
  booktitle={Proceedings of the 2021 Edition of the Automatic Speaker Verification and Spoofing Countermeasures Challenge},
  year={2021},
  pages={47--54}
}

@misc{diffssd2024,
  author={Bhagtani, Kratika and Yadav, Amit Kumar Singh and Bestagini, Paolo and Delp, Edward J.},
  title={DiffSSD: A Diffusion-Based Dataset For Speech Forensics},
  year={2024},
  eprint={2409.13049},
  archivePrefix={arXiv},
  primaryClass={cs.SD}
}

@inproceedings{partialspoof2021,
  author={Zhang, Lin and Wang, Xin and Cooper, Erica and Yamagishi, Junichi and Patino, Jose and Evans, Nicholas},
  title={An Initial Investigation for Detecting Partially Spoofed Audio},
  booktitle={Interspeech 2021},
  pages={1654--1658},
  year={2021},
  doi={10.21437/Interspeech.2021-1001}
}

@article{partialspoof2023,
  author={Zhang, Lin and Wang, Xin and Cooper, Erica and Evans, Nicholas and Yamagishi, Junichi},
  title={The PartialSpoof Database and Countermeasures for the Detection of Short Fake Speech Segments Embedded in an Utterance},
  journal={IEEE/ACM Transactions on Audio, Speech, and Language Processing},
  volume={31},
  pages={813--825},
  year={2023},
  doi={10.1109/TASLP.2022.3233236}
}

@inproceedings{xlsr2022,
  author={Babu, Arun and Wang, Changhan and Tjandra, Andros and Lakhotia, Kushal and Xu, Qiantong and Goyal, Naman and Singh, Kritika and von Platen, Patrick and Saraf, Yatharth and Pino, Juan and Baevski, Alexei and Conneau, Alexis and Auli, Michael},
  title={XLS-R: Self-Supervised Cross-Lingual Speech Representation Learning at Scale},
  booktitle={Interspeech 2022},
  pages={2278--2282},
  year={2022},
  doi={10.21437/Interspeech.2022-143}
}

@misc{trace2026,
  author={Khan, Awais and Farooq, Muhammad Umar and Uddin, Kutub and Malik, Khalid},
  title={TRACE: Training-Free Partial Audio Deepfake Detection via Embedding Trajectory Analysis of Speech Foundation Models},
  year={2026},
  eprint={2604.01083},
  archivePrefix={arXiv},
  primaryClass={eess.AS}
}

@inproceedings{tdl,
  author={Xie, Yiting and Cheng, Hong and Wang, Yuxuan and Ye, Lin},
  title={An Efficient Temporary Deepfake Location Approach Based Embeddings for Partially Spoofed Audio Detection},
  booktitle={ICASSP 2024},
  pages={966--970},
  year={2024}
}

@inproceedings{cfprf,
  author={Wu, Jing and Lu, Wenjun and Luo, Xiaolong and Yang, Rui and Wang, Qi and Cao, Xiaochun},
  title={Coarse-to-Fine Proposal Refinement Framework for Audio Temporal Forgery Detection and Localization},
  booktitle={Proceedings of the 32nd ACM International Conference on Multimedia},
  pages={7395--7403},
  year={2024},
  doi={10.1145/3664647.3681478}
}

@inproceedings{bam,
  author={Zhong, Jiaming and Li, Baoyuan and Yi, Jiangyan},
  title={Enhancing Partially Spoofed Audio Localization with Boundary-Aware Attention Mechanism},
  booktitle={Interspeech 2024},
  pages={4838--4842},
  year={2024},
  doi={10.21437/Interspeech.2024-1477}
}

@inproceedings{ago,
  author={Zeng, Siding and Yi, Jiangyan and Tao, Jianhua and He, Jiayi and Lian, Zheng and Liang, Shuang and Zhang, Chu and Chen, Yixuan and Zhang, Xiaohui},
  title={Adversarial Training and Gradient Optimization for Partially Deepfake Audio Localization},
  booktitle={ICASSP 2025},
  pages={1--5},
  year={2025}
}

@inproceedings{gncl,
  author={Ge, Zhiyong and Xu, Xin and Guo, Hao and Yang, Zijiang and Schuller, Bj{\"o}rn},
  title={GNCL: A Graph Neural Network with Consistency Loss for Segment-Level Spoofed Speech Detection},
  booktitle={ICASSP 2025},
  pages={1--5},
  year={2025}
}

@misc{cqcc_lcnn,
  author={Todisco, Massimiliano and Delgado, H{\'e}ctor and Evans, Nicholas},
  title={Constant Q Cepstral Coefficients and Light CNN Baseline for Spoofing Countermeasures},
  year={2016},
  note={CQCC-based baseline cited for PartialSpoof comparison}
}

@inproceedings{lcnn_lfcc,
  author={Zhang, Lin and Wang, Xin and Cooper, Erica and Yamagishi, Junichi},
  title={Multitask Learning in Utterance-Level and Segmental-Level Spoof Detection},
  booktitle={Proceedings of the ASVspoof 2021 Workshop},
  year={2021}
}

@inproceedings{lcnn_w2v2,
  author={Zhang, Lin and Wang, Xin and Cooper, Erica and Yamagishi, Junichi},
  title={Multitask Learning in Utterance-Level and Segmental-Level Spoof Detection},
  booktitle={Proceedings of the ASVspoof 2021 Workshop},
  year={2021}
}

@inproceedings{selcnn,
  author={Wu, Haibin and Kuo, Hsin-Chieh and Zheng, Nanxin and Hung, Kuo-Hsuan and Lee, Hung-Yi and Tsao, Yu and Wang, Hsin-Min and Meng, Helen},
  title={Partially Fake Audio Detection by Self-Attention-Based Fake Span Discovery},
  booktitle={ICASSP 2022},
  pages={9236--9240},
  year={2022}
}

@misc{w2v2_mlp,
  author={Tak, Hyeji and Todisco, Massimiliano and Wang, Xin and Jung, Jee-weon and Yamagishi, Junichi and Evans, Nicholas},
  title={Automatic Speaker Verification Spoofing and Deepfake Detection Using wav2vec 2.0 and Data Augmentation},
  year={2022},
  eprint={2202.12233},
  archivePrefix={arXiv},
  primaryClass={eess.AS}
}

@inproceedings{wu2026wildspoof,
  title={WildSpoof: advancing in-the-wild data in Text-to-Speech generation and Spoofing-aware automatic speaker verification},
  author={Wu, Yihan and Jung, Jee-Weon and Shim, Hye-Jin and Cheng, Xin and Wang, Xin},
  booktitle={ICASSP 2026-2026 IEEE International Conference on Acoustics, Speech and Signal Processing (ICASSP)},
  pages={21922--21924},
  year={2026},
  organization={IEEE}
}

@inproceedings{kinnunen2012vulnerability,
  title={Vulnerability of speaker verification systems against voice conversion spoofing attacks: The case of telephone speech},
  author={Kinnunen, Tomi and Wu, Zhi-Zheng and Lee, Kong Aik and Sedlak, Filip and Chng, Eng Siong and Li, Haizhou},
  booktitle={2012 IEEE international conference on acoustics, speech and signal processing (ICASSP)},
  pages={4401--4404},
  year={2012},
  organization={IEEE}
}

@article{li2025hq,
  title={HQ-MPSD: A Multilingual Artifact-Controlled Benchmark for Partial Deepfake Speech Detection},
  author={Li, Menglu and Alber, Majd and Asgarianamiri, Ramtin and Zhao, Lian and Zhang, Xiao-Ping},
  journal={arXiv preprint arXiv:2512.13012},
  year={2025}
}

@inproceedings{ravanelli2018speaker,
  title={Speaker recognition from raw waveform with sincnet},
  author={Ravanelli, Mirco and Bengio, Yoshua},
  booktitle={2018 IEEE spoken language technology workshop (SLT)},
  pages={1021--1028},
  year={2018},
  organization={IEEE}
}

@article{fei2026multi,
  title={Multi-scale Temporal Causal Network for Speech Emotion Recognition},
  author={Fei, Sijia and Feng, Qiang and Gao, Fei},
  journal={IEEE Access},
  year={2026},
  publisher={IEEE}
}

@inproceedings{tran2024parallelchain,
  title={ParallelChain Lab’s antispoofing systems for ASVspoof 5},
  author={Tran, Thien and Bui, Thanh Duc and Simatis, Panagiotis},
  booktitle={The automatic speaker verification spoofing countermeasures workshop (ASVspoof 2024). ISCA, Kos, Greece},
  pages={9--15},
  year={2024}
}

% \section{Open Science}
% To support reproducibility, we have provided an anonymized artifact containing the full PyTorch implementation of Adv-TGD. The repository includes scripts for SGSM generation, LoRA optimization.
% The artifact is available at: \url{https://anonymous.4open.science/r/Adv-TGD-Artifact-38E4/README.md}

%%
%% If your work has an appendix, this is the place to put it.
% \appendix

% \section{appendix}

\end{document}